\documentclass[conference]{IEEEtran}
\IEEEoverridecommandlockouts
\usepackage{cite}
\usepackage{amsmath,amssymb,amsfonts}
\usepackage{algorithmic}
\usepackage{graphicx}
\usepackage{textcomp}
\usepackage{xcolor}
\usepackage{balance}
\usepackage[authormarkup=none]{changes}
\usepackage{enumerate}
\usepackage{multirow}
\usepackage{algorithm}
\usepackage{algorithmic}
\usepackage{subfigure}
\usepackage{subcaption}
\usepackage{pifont}

\graphicspath{{figures/}}
\newcommand{\tabref}[1]{Table~\ref{#1}}
\newcommand{\figref}[1]{Figure~\ref{#1}}
\newcommand{\secref}[1]{Section~\ref{#1}}

\newcommand{\CASE}[1]{\STATE \textbf{case} #1\textbf{:} \begin{ALC@g}}
\newcommand{\ENDCASE}{\end{ALC@g}}

\newcommand{\DEFAULT}{\STATE \textbf{default:} \begin{ALC@g}}
\newcommand{\ENDDEFAULT}{\end{ALC@g}}
\newcommand{\DEFAULTLINE}[1]{\STATE \textbf{default:} }

\def\BibTeX{{\rm B\kern-.05em{\sc i\kern-.025em b}\kern-.08em
    T\kern-.1667em\lower.7ex\hbox{E}\kern-.125emX}}
\begin{document}

% \title{Conference Paper Title*\\
% {\footnotesize \textsuperscript{*}Note: Sub-titles are not captured in Xplore and
% should not be used}
% \thanks{Identify applicable funding agency here. If none, delete this.}
% }
\title{Shylock: Causal Discovery in Multivariate Time Series based on Hybrid Constraints}

% \author{\IEEEauthorblockN{Anonymous Author(s)}}
\author{\IEEEauthorblockN{Shuo Li, Keqin Xu\thanks{Shuo Li and Keqin Xu contribute equally.}}
\IEEEauthorblockA{\textit{Institute of Software,Chinese Academy of Sciences}\\ 
\textit{University of Chinese Academy of Sciences}\\ \textit{Beijing 100049, China} \\
\{lishuo,xukeqin\}19@otcaix.iscas.ac.cn}
\and
\IEEEauthorblockN{Jie Liu*, Dan Ye \thanks{Jie Liu is corresponding author.}}
\IEEEauthorblockA{\textit{Institute of Software,Chinese Academy of Sciences}\\ 
\textit{University of Chinese Academy of Sciences}\\ \textit{Beijing 100049, China} \\
\textit{University of Chinese Academy of Sciences}\\
\textit{Nanjing, Nanjing 211135, China}\\
\{ljie,yedan\}@otcaix.iscas.ac.cn}
% \and
% \IEEEauthorblockN{Lin Xiong }
% \IEEEauthorblockA{\textit{China International Engineering Consulting Corporation} \\
% Beijing, China \\
% xl@ciecc.com.cn}
% \and
% \IEEEauthorblockN{Dan Ye}
% \IEEEauthorblockA{\textit{Institute of Software,Chinese Academy of Sciences}\\ 
% \textit{University of Chinese Academy of Sciences}\\ \textit{Beijing 100049, China}  \\
% yedan@otcaix.iscas.ac.cn}
}

\maketitle

\begin{abstract}
Causal relationship discovery has been drawing increasing attention due to its prevalent application. Existing methods rely on human experience, statistical methods, or graphical criteria methods which are error-prone, stuck at the idealized assumption, and rely on a huge amount of data. And there is also a serious data gap in accessing Multivariate time series(MTS) in many areas, adding difficulty in finding their causal relationship. Existing methods are easy to be over-fitting on them.

To fill the gap we mentioned above, in this paper, we propose \textbf{Shylock}, a novel method that can work well in both few-shot and normal MTS to find the causal relationship. 
Shylock can reduce the number of parameters exponentially by using group dilated convolution and a sharing kernel, but still learn a better representation of variables with time delay. 
By combing the global constraint and the local constraint, Shylock achieves information sharing among networks to help improve the accuracy. 
To evaluate the performance of Shylock, we also design a data generation method to generate MTS with time delay. We evaluate it on commonly used benchmarks and generated datasets.
Extensive experiments show that Shylock outperforms two existing state-of-art methods on both few-shot and normal MTS.
We also developed Tcausal, a library for easy use and deployed it on the earthDataMiner platform \footnote{A Cloud-Based Big Earth Data Intelligence Analysis Platform.}.
\end{abstract}

\begin{IEEEkeywords}
Multivariate time series, Causality, Causal discovery
\end{IEEEkeywords}

\section{Introduction}

Time series data can help uncover relationships between variables. Multivariate time series (MTS) data is generated when recording time series from a wide range of sensors.
% In MTS data, there contains more than one time-dependent variable, and each variable depends not only on historical values but may also depend on other variables.

Existing researches utilize a huge amount of MTS for forecasting. It has seen tremendous applications in the domains of economics, finance, bioinformatics, and traffic \cite{wu2020connecting} \cite{cui2021metro}. 
% However, these relationships do not represent the causal relationship among them. 

But in recent years, some researchers are more concerned with the causal relationships among the variables in MTS data. By identifying causality, researchers and practitioners can gain a deeper understanding of how changes in one variable affect other variables, and can make more informed decisions and predictions.
% A few works have been proposed for causal discovery in MTS data. 
For example, in recent years, the rapidly treated Arctic sea ice has attracted much attention which is also an important point in the global Sustainable Development Goals which lay out a comprehensive and ambitious agenda for global development\footnote{https://sdgs.un.org/}. 
Knowing the causal relationship between retreated Arctic sea ice with other factors can further help protect the environment. 
So some researchers struggled to collect these related data, such as the global land degradation rate and the world's groundwater usage rate, to find their causal relationships. 
These data are extremely difficult to collect \footnote{https://blogs.worldbank.org/opendata/are-we-there-yet-many-countries-dont-report-progress-all-sdgs-according-world-banks-new}. 
We refer to these time series with a tiny amount of data as few-shot multivariate time series.
Furthermore, there usually exists a time delay among variables in MTS data, which further increases the difficulty of finding causal relationships.

Traditional methods rely on human experience to find causal relations, which is time-consuming and error-prone. Some statistical methods \cite{Granger2001} \cite{Geweke1982} \cite{chen2004analyzing} \cite{LuoLKC15} or graphical criteria methods rely on a non-trivial combination of probability axioms\cite{MatrixEquations} \cite{NoTears}.
% to find causal relationships over multivariate data. 
However, these methods can not work well on MTS data as they can not characterize time delay features of data or are stuck at idealized assumptions (e.g., the data is without any noise).  Causal discovery aims to discover direct cause-effect relationships for both instantaneous and delayed causes\cite{huang2021benchmarking}.
% For example, NOTEARS \cite{NoTears} formalizes the causal discovery as a continuous optimization problem but does not solve the time delay problem of variables. Its variant, NTS-NOTEARS \cite{ntsNOTEARS}, extends NOTEARS with optimized constraints to work on multivariate time series, but the constraints are difficult to actualize \cite{ng2022convergence}. 

In recent years, benefiting from the development of deep learning techniques, some of the advanced deep learning-based methods, such as TCDF \cite{tcdf}, have been proposed to find causal relationships. It can model the time delay by the neural networks to enhance the ability to learn causal relationships. 
However, these methods depend on a huge amount of data and parameters. In many areas, it may be difficult and infeasible to collect a large amount of data, and the data scale is so small even less than hundreds of items. There are serious data gaps in assessing the aforementioned data. Furthermore, these methods suffer from being over-parametrization and difficult to converge, and difficult to learn the generalized temporal feature expression when applied on few-shot MTS. At the same time, few-shot MTS is always characterized by high noise and time delay, increasing the difficulty of finding the causal relationship. For example, the total number of parameters in TCDF is not less than $O (nnf)$, where $n$ represents the num of time series (also called variables) and $f$ represents the
% length of variables(also refers to the length of timesteps). 
length of receptive field(almost the same as the length of timesteps). 
When applied TCDF on few-shot MTS, our experiment proves that it is over-fitting. Detailed information is shown in \secref{fit_ability1}.

% In order to overcome the existing problems on the few-shot MTS dataset
To fill the gap we mentioned above, we propose \textbf{Shylock}, a novel method that can effectively find causal relationships on multivariate time series even on few-shot multivariate time series.
% to find the \textbf{c}ausa\textbf{l} r\textbf{e}lationship 
% for \textbf{f}ew-shot multivariate \textbf{t}ime series.
Shylock independently models the causal relationship between variables using a neural network. To solve the time delay of variables and reduce the number of parameters exponentially, Shylock uses group dilated convolution in each network to learn a better representation of variables, and a sharing kernel to learn local causal relationships among variables. Then Shylock uses a global loss to obtain the global causal relationships, which are represented by the attention matrix. To identify cyclic causal relationships. Shylock conducts constraints on the global loss and attention and eliminates cyclic causal relationships by DAG. 

To evaluate Shylock, we design a lightweight method to generate MTS data with time delay. Based on that, we generate few shot datasets and use Shylock to find causal relationships among the variables. Besides, we also choose a benchmark, FMRI \cite{friston2009causal}, to evaluate the performance of Shylock. The experiment results show that Shylock is effective and efficient in finding causal relationships in MTS data, and significantly outperforms two existing state-of-art methods, NOTEARS and TCDF. Overall, our contributions are as follows:

% Then, by combining the local constraints of each 
% sub-network and global constraints to filter out 
% the false causal relationship and get the true relationship. 

\begin{itemize}
  \item To the best of our knowledge, we are the first to emphasize the causal relationship discovery on few-shot multivariate time series. We propose \textbf{Shylock}, a neural-network-based method that incorporates hybrid constraints to mine causal relationships on Multivariate Time Series even  in limited data scenarios.
  % \textbf{c}ausa\textbf{l}  r\textbf{e}lationship for \textbf{f}ew-shot 
  % Multivariate \textbf{t}ime Series. 
  
  \item SShylock reduces the number of parameters exponentially and uses hybrid constraints to facilitate information sharing during training and prediction without compromising performance. It addresses time delays in MTS data and minimizes parameterization by employing group dilated convolutions and a shared kernel to learn local causal relationships. Shylock further applies global constraints via a DAG to eliminate cyclic causal relationships, combining local and global constraints to infer causal connections.
    
  \item To evaluate the effectiveness of Shylock, we developed a lightweight method to generate MTS data with time delays. Experiments on the generated datasets and a common benchmark show that Shylock outperforms two state-of-the-art methods in identifying causal relationships in MTS data with time delays.
  % a lightweight implementation of the method is presented. We propose a method to generate few-shot multivariate time series, similar to NOTEARS\cite{NoTears} but making the generated data with time delay.
%   To promote replication and open science, the source code and datasets used in this paper are available. \lsadd{ at http:} We also released a third-party library \textbf{Tcausal}, which has been deployed on earthDataMiner for easy use. 
  
\end{itemize}

% \section{Motivation}
% pass 

\section{Introduction}

Time series data can help uncover relationships between variables. Multivariate time series (MTS) data is generated when recording time series from a wide range of sensors.
% In MTS data, there contains more than one time-dependent variable, and each variable depends not only on historical values but may also depend on other variables.

Existing researches utilize a huge amount of MTS for forecasting. It has seen tremendous applications in the domains of economics, finance, bioinformatics, and traffic \cite{wu2020connecting} \cite{cui2021metro}. 
% However, these relationships do not represent the causal relationship among them. 

But in recent years, some researchers are more concerned with the causal relationships among the variables in MTS data. By identifying causality, researchers and practitioners can gain a deeper understanding of how changes in one variable affect other variables, and can make more informed decisions and predictions.
% A few works have been proposed for causal discovery in MTS data. 
For example, in recent years, the rapidly treated Arctic sea ice has attracted much attention which is also an important point in the global Sustainable Development Goals which lay out a comprehensive and ambitious agenda for global development\footnote{https://sdgs.un.org/}. 
Knowing the causal relationship between retreated Arctic sea ice with other factors can further help protect the environment. 
So some researchers struggled to collect these related data, such as the global land degradation rate and the world's groundwater usage rate, to find their causal relationships. 
These data are extremely difficult to collect \footnote{https://blogs.worldbank.org/opendata/are-we-there-yet-many-countries-dont-report-progress-all-sdgs-according-world-banks-new}. 
We refer to these time series with a tiny amount of data as few-shot multivariate time series.
Furthermore, there usually exists a time delay among variables in MTS data, which further increases the difficulty of finding causal relationships.

Traditional methods for causal discovery depend heavily on human expertise, which is time-consuming and error-prone. Statistical techniques \cite{Granger2001, Geweke1982, chen2004analyzing, LuoLKC15} and graphical methods based on probability axioms \cite{MatrixEquations, NoTears} are often employed for multivariate data. However, they struggle with MTS data due to limitations in capturing time delays or reliance on unrealistic assumptions, such as noise-free data \cite{huang2021benchmarking}. 
% Causal discovery aims to discover direct cause-effect relationships for both instantaneous and delayed causes\cite{huang2021benchmarking}.
% For example, NOTEARS \cite{NoTears} formalizes the causal discovery as a continuous optimization problem but does not solve the time delay problem of variables. Its variant, NTS-NOTEARS \cite{ntsNOTEARS}, extends NOTEARS with optimized constraints to work on multivariate time series, but the constraints are difficult to actualize \cite{ng2022convergence}. 

Benefiting from the development of deep learning techniques, some of the advanced deep learning-based methods, such as TCDF \cite{tcdf}, have shown promise in modeling time delays and causal discovery. 
% It can model the time delay by the neural networks to enhance the ability to learn causal relationships. 
However, these methods require large amounts of data and parameters, which can be difficult to obtain, especially in fields with small datasets (often fewer than hundreds of samples).
Furthermore, these techniques suffer from over-parameterization, poor convergence, and challenges in learning generalized temporal features, particularly when applied to few-shot MTS data, which is often noisy and involves time delays. 
% At the same time, few-shot MTS is always characterized by high noise and time delay, increasing the difficulty of finding the causal relationship. 
For instance, TCDF has at least $O(nnf)$ parameters, where $n$ is the number of time series and $f$ is the receptive field length (approximately equal to the number of timesteps). Our experiments show that TCDF tends to overfit when applied to few-shot MTS, as detailed in \secref{fit_ability1}.

To fill the gap we mentioned above, we propose \textbf{Shylock}, a novel method designed to effectively discover causal relationships in multivariate time series (MTS), even in few-shot settings. Shylock independently models causal relationships between variables using neural networks. To handle time delays and reduce parameter counts exponentially, Shylock employs group dilated convolutions within each network to learn improved representations of variables, while a shared kernel captures local causal relationships. It then uses a global loss to infer global causal relationships, represented by the attention matrix. To address cyclic causal relationships, Shylock introduces constraints on the global loss and attention, eliminating cycles through a directed acyclic graph (DAG). 

To evaluate Shylock, we design a To evaluate Shylock, we developed a lightweight method to generate MTS data with time delays, creating few-shot datasets for causal discovery. We also benchmarked Shylock against FMRI data \cite{friston2009causal} to assess its performance. Experimental results demonstrate that Shylock is both effective and efficient in identifying causal relationships in MTS data, significantly outperforming two state-of-the-art methods, NOTEARS and TCDF. Overall, our contributions are as follows:

% Then, by combining the local constraints of each 
% sub-network and global constraints to filter out 
% the false causal relationship and get the true relationship. 

\begin{itemize}
  \item To the best of our knowledge, we are the first to emphasize the causal relationship discovery on few-shot multivariate time series. We propose \textbf{Shylock}, a neural-network-based method that incorporates hybrid constraints to mine causal relationships on Multivariate Time Series even in the case of few-shot series.
  % \textbf{c}ausa\textbf{l}  r\textbf{e}lationship for \textbf{f}ew-shot 
  % Multivariate \textbf{t}ime Series. 
  
  \item Shylock can reduces the number of parameters exponentially and leverages hybrid constraints to facilitate information sharing during training and prediction without degrading the model performance . 
  To solve time delay in MTS data and release parametrization, Shylock employs group dilated convolutions and a sharing kernel to learn local causal relationships among variables. 
  Shylock conducts global constraints to identify and eliminate cyclic causal relationships using DAG. 
  By combining the local and global constraints, Shylock imposes a global loss to find causal relationships.
    
  \item To evaluate the effectiveness of Shylock, we also proposed a lightweight method to generate MTS data with time delay. Based on the generated dataset and commonly used benchmark, experiment results show that, compared with two existing state-of-art methods, Shylock is more effective and efficient in finding causal relationships among MTS data with time delay.
  % a lightweight implementation of the method is presented. We propose a method to generate few-shot multivariate time series, similar to NOTEARS\cite{NoTears} but making the generated data with time delay.
%   To promote replication and open science, the source code and datasets used in this paper are available. \lsadd{ at http:} We also released a third-party library \textbf{Tcausal}, which has been deployed on earthDataMiner for easy use. 
  
\end{itemize}

% \section{Motivation}
% pass 

\section{problem statement}
\begin{figure}[h]
  \centering
  \includegraphics[scale=0.35]{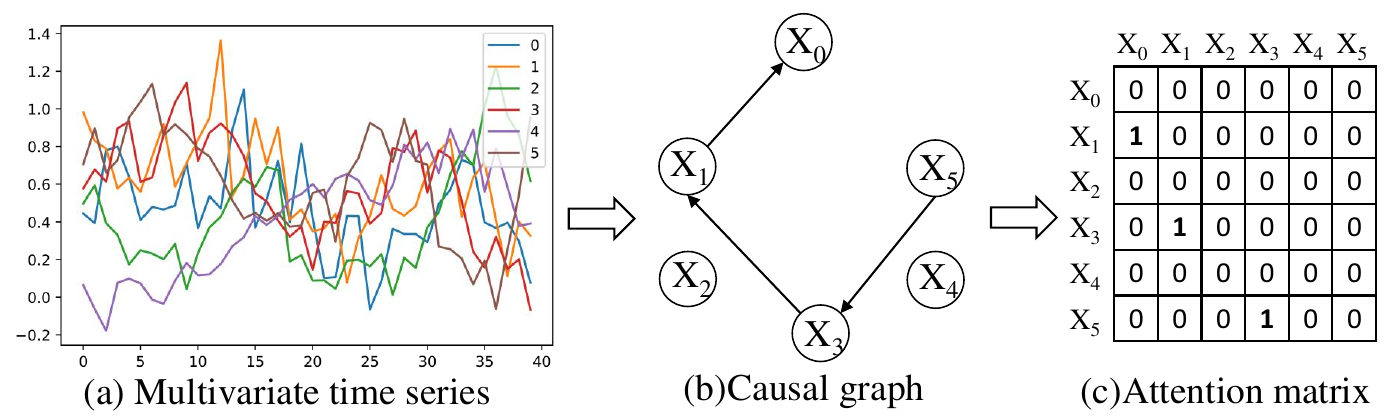}
  % \caption{The format conversion between few-shot multivariate 
      %  time series, causal graph, and attention matrix.}
  \caption{Temporal causal graph representation.}
  \label{data_relation}
  \vspace{-10px}
  \end{figure}

% \subsection{Problem Definition}
\textbf{Multivariate Time Series data:} 
Temporal causal discovery between multivariate time series from 
observational can be formulated as follows: 

As shown in \figref{data_relation}(a), the dataset $X = \left\{ x_1,...,x_n \right\}$ are consisted of $n$ observed time series of the same length $t$. These time series are also called variables. For sake of brevity, in the following paper, we denoted the times series as variables in the following, and data collected by each observation variable can be aligned in time.
% There is also a special causal relationship, 
% self-causation, which means the variable is both the cause and the effect.
% time delay definition

\textbf{Causal graph:}
Then these causal relationships are commonly
expressed as a causal graph $\mathcal{G}$, which can be represented as a directed acyclic graph (DAG) according to the common assumption. As shown in \figref{data_relation}(b), the vertices are time series $x_k \in X $ and the arrows are direct causal relationships. 

\textbf{Attention matrix:}In order to formalize the graphical constraints on DAG and for the convenience of calculations, 
an attention matrix $A$ is introduced, shown in \figref{data_relation}(c). $a_{i,j}>= threshold$ represents the $i_{th}$ time series $x_i$ as the effect and the $j_{th}$ time series $x_j$ as the cause and conversely.

\begin{figure*}[h]
  \centering
  \includegraphics[width=0.7\linewidth, ]{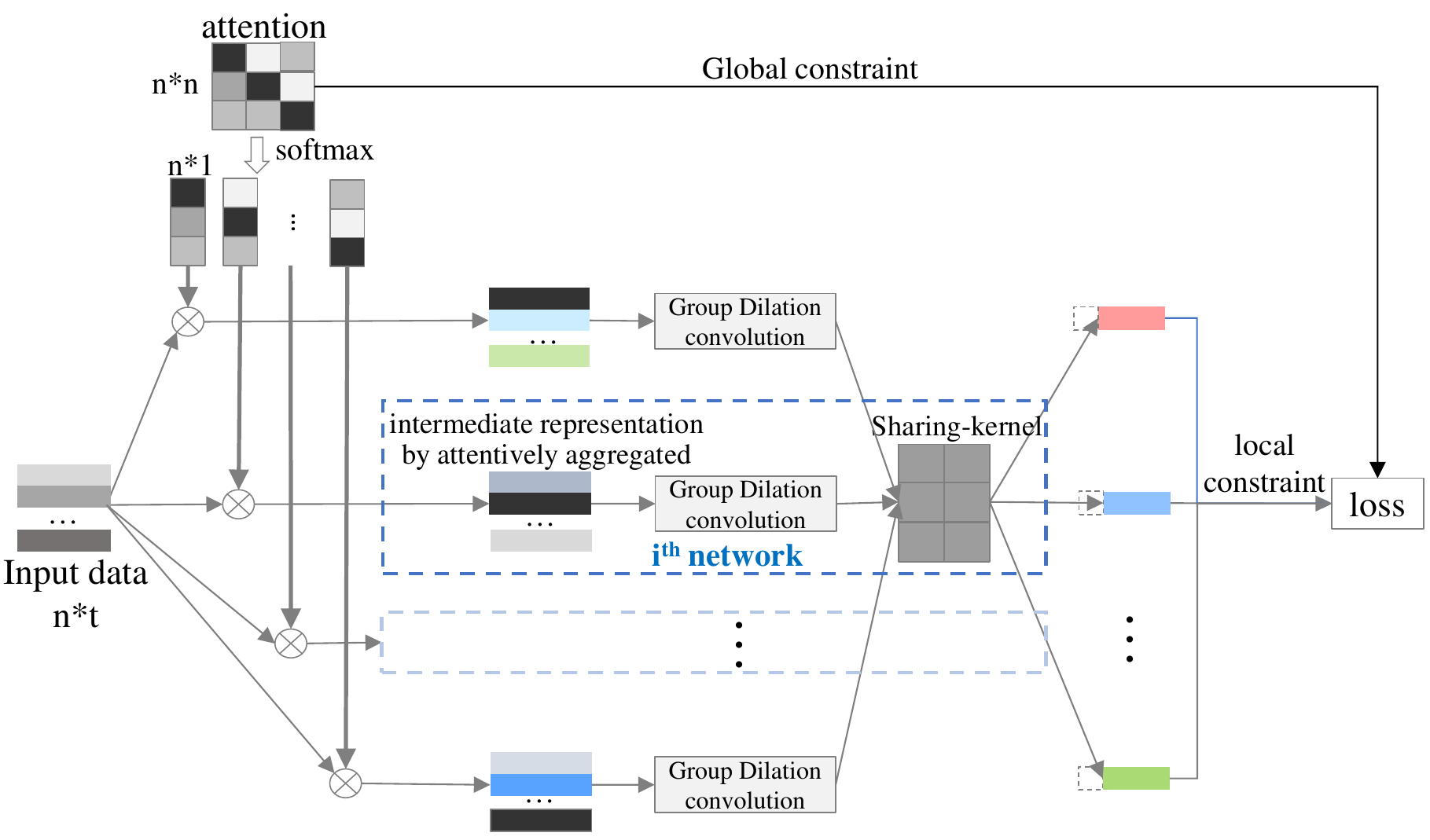}
  \caption{Approach Overview of Shylock.}
  \label{architecture}
  % \vspace{-10px}
  \end{figure*}

\section{Shylock}
This section presents a neural network-based approach for mining causal relationships in multivariate time series (MTS). \figref{architecture} illustrates the architecture of Shylock, comprising two key steps:

\begin{enumerate}[(1)]
  \item \textbf{Attention-based local causal discovery:} 
  Shylock models potential causes for each variable by constructing $n$ sub-convolutional neural networks (CNNs), each targeting a single variable. To address overfitting on few-shot MTS, shared convolution kernels minimize parameter size, while a grouped dilated convolution module captures time-delayed causal effects.
  \item \textbf{Adjacent matrix constraint-based global causal discovery:} 
  By combining global constraints with local fitting objectives, Shylock ensures acyclic causal relationships. Attention vectors from sub-CNNs are combined to form an attention matrix, but due to the lack of direct information sharing, the matrix alone does not guarantee acyclicity. Therefore, adjacency matrix constraints are applied to enforce this property.
\end{enumerate}

In the following sections, we describe each step in more detail.
For easy description, we formalize some related definitions and describe each step in detail in the following sections.

\subsection{Attention-based local causal discovery with Parameter Sharing} 
Shylock identifies causal relationships by building CNNs for each variable $x_i$,as shown in the blue dotted box of \figref{architecture}. The CNNs use two convolution kernel types: \ding{172}Grouped Dilated Convolution, which Models time-series features.
\ding{173}One-dimensional Convolution, which aggregates these features to capture potential causes for the target variable.

\textbf{Attention matrix:} As shown on \figref{architecture}
The matrix $A$ represents attention relationships between variables, defined as $A = \left\{ a_{1},...,a_{n} \right\}$, 
where $a_{k}= \left\{ a_{1,k},...,a_{n,k} \right\}$ is an $N*1$ vector representing weights for the $k_{th}$ network. 
Initially, self-causation weights $a_{kk}$ are preset to $\alpha$ (commonly 0), with others set to 1. 
During training, the attention matrix adjusts dynamically, ultimately determining the causal relationships based on thresholded attention weights $a_{i,j}$. Further details on the global constraints applied to these networks are discussed in \secref{sec:adj_global}.

\begin{figure}[h]
  \centering
  \includegraphics[scale=0.3]{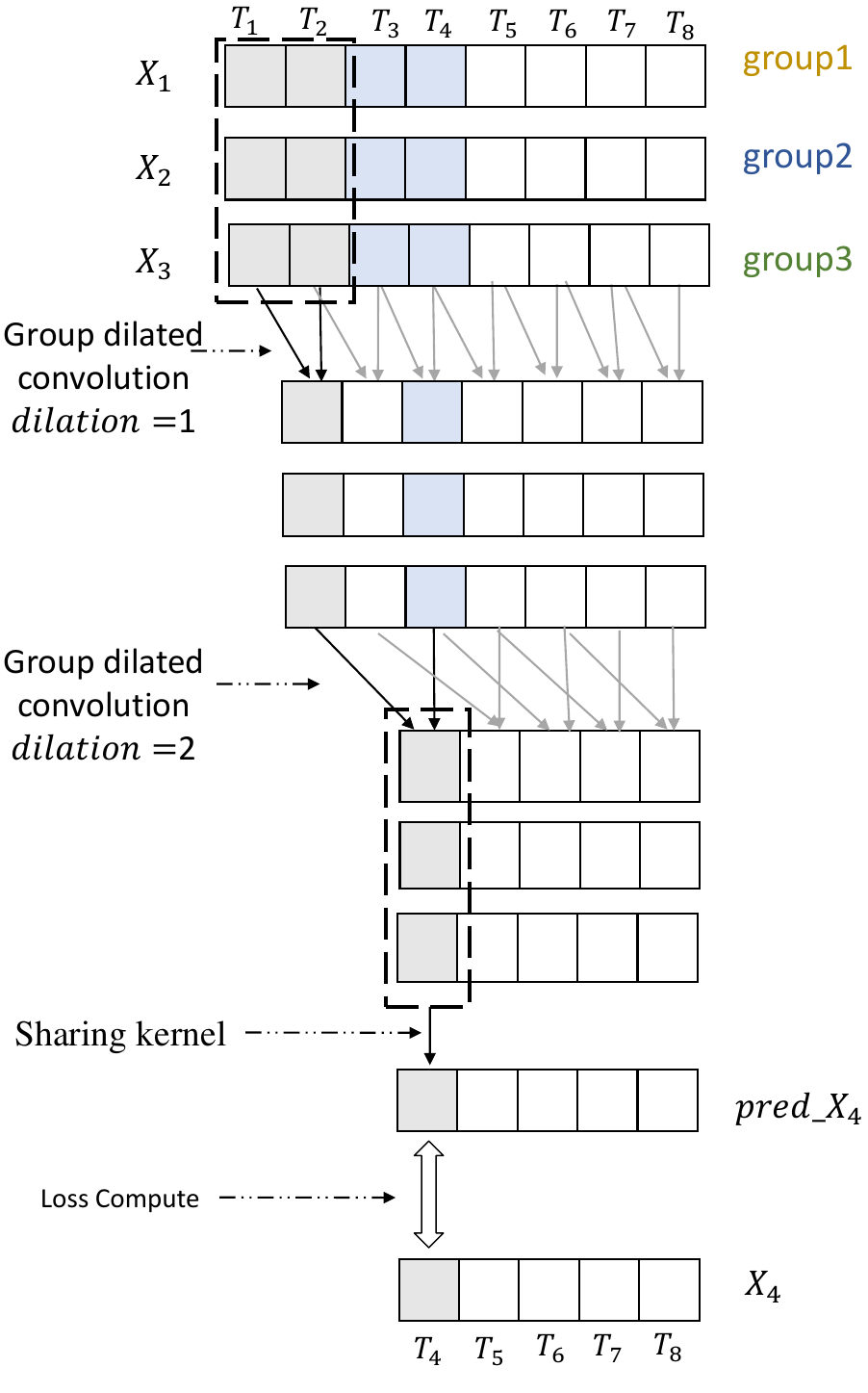}
  \caption{Group dilated convolution.}
  \label{Group_dilated_convolution}
  \vspace{-10px}
\end{figure}

\textbf{Group dilated convolution:}
Each network aims to discover causal relationships between a target variable and others. Sparse MTS often exhibit long time delays, necessitating a receptive field larger than the maximum lag $K$ for accurate modeling. Existing methods often require convolution kernels with $N*K$ parameters, leading to overfitting. To balance receptive field size and parameter efficiency, we adopt group dilated convolution for univariate time-series modeling (\figref{Group_dilated_convolution}). 

Specifically, our method employs $N$ sets of convolution filter groups to model feature representations. To capture temporal features effectively without loss of resolution, dilated convolutions are utilized, allowing exponential expansion of the receptive field \cite{8941073}. With a dilation factor of length $d_{l}$, the receptive filed reaches approximately $d_{l} = T^{l}-1$, mitigating long time delays with only $O(LNT)$ parameters. For example, as shown in \figref{Group_dilated_convolution}, four variables $X_1, X_2, X_3, X_4$ each with eight timesteps benefit from this approach.

\textbf{Sharing kernel:} 
To model variable relationships while avoiding overfitting, shared kernels are introduced. These kernels reduce interference and parameter count by focusing on local variable associations. In the $i_{th}$ network, shared kernels model the filtered representation of the $i_{th}$ variable after group dilated convolution, updating the attention matrix via loss computation.
For variables $Q$, $K$, and $V$ with a causal relationship $Q \rightarrow K$ (where $Q$ represents the cause, $k$ represents the effect, and $V$ represents no causal relationship. So, during the causal discovery among these variables, $V$ may introduce noise. 
We use $W = \left\{ w_{1},...,w_{n} \right\}$ to represent the weight 
parameter of causal effect, and $\delta$ to represent the noise. 
They can be formalized as:
\begin{equation}
  K = WQ + \delta , w_{i\notin q} = 0
\end{equation}
Therefore, the inference of causal relationships between variables does not require the participation of all data.
At the same time, it's more likely that there will be a consistent 
set of variables depending on different variables on few-shot MTS. 
For all networks, the sparse model of the convolution kernel is 
different for the parameters of the convolution kernel only act on 
the association inference between local variables.
So we propose sharing the kernel for the second kind 
of the kernel to reduce the parameters and enhance anti-interference ability.

\subsection{DAG constraint-based global causal discovery:}
\label{sec:adj_global}
Each attention-based neural network models the causal relationship for a variable as the cause and other variables as the effect. However, these networks independently focus on individual variables and cannot optimize from a global perspective to eliminate cyclic relationships. Directed acyclic graphs (DAGs) provide a strict acyclicity constraint, making them suitable for causal discovery.
We formalize the causal relationship as 
$x_{i} = w_{i}X + \delta_{i}$,  with attention matrix $A$ representing causal relationships. If $W_{i,j} > \text{threshold}$, we define $A_{i,j} = 1$ to indicate causality. We impose a global penalty on $W$ via $h(W)$, formulated as:
\begin{equation}
  h(W) = tr(\sum_{k=1} A^{k})
\end{equation}   
This function ensures acyclic relationships when $h(W) = 0$, and its value increases with the presence of loops. $h(W)$ also guarantees numerical stability for function and gradient evaluations.
Suppose the adjacent matrix of directed acyclic garph $G$ as $A$, then 
$a_{ij}^{(1)}=1$ in $A^{1}$ represents there is a path from 
$i_{th}$ variable to the $j_{th}$ variable. 
% If $a_{ij}^{(k)}=1$ 
% in $A^{k}$ is more than 0, it represents there is a path of 
% length $k$ from $i_{th}$ variable to the $j_{th}$ variable.
Then $A^2 = A^1*A$, which can be further represented as 
$a_{ij}^{(2)}=\sum_{p=0}^n a_{ip}^{(1)} a_{pj} $. If $a_{ij}^{(2)}>0$, 
there is a 2-length path from $V_{i} \to V_{j} $. 
And so on, if $a_{ij}^{(k)}$ in $A^{(1)}$ is larger than 0, there is 
a $k$-length path from $v_{i}$ to $v_{j} $. Then we can deduce that
% \begin{equation}
  $tr(A^{k}) = \sum_{i=0}^{n} a_{ii}^{k}, k>0$
% \end{equation}  
. $tr(A^{k})>0$ represents that there is a $k$-length path from
$i_{th}$ variable to itself. To reduce it, the coefficient $h(W)$ is applied to punish the causal loops of different lengths, denoted as:
$h(W) = tr(\sum_{k=1} \beta^{k}A^{k})$
Based on this, the whole loss of our method is: 
\begin{equation}
  \begin{aligned}
    \ell(\cdot) &= \sum_{k=0}^{n-1} (k+1) W^{k} \\
                &=\sum_{k=0}^{m} \frac{1}{n} \sum_{i=0}^{n}(y_{i}-f(x_{i}|W_{a}^{k}, W_{c}))^{2} +\alpha tr(\sum_{l=1}^{m}W_{a}^{l}) + \beta |W_{c}| 
  \end{aligned}
  \label{global_loss}
\end{equation}  
where we equally treat circular causal relationships of different lengths
\eqref{global_loss} can help transmit the global loss into local loss
for each network. 
In this method, the discrete networks aiming at each variable 
are recombined into a global continuous optimization model, 
and the combination of local single-objective high-precision causal discovery and global causal graph constraint is realized.

\section{EXPERIMENTS AND RESULTS}
In this section, we apply Shylock to three benchmarks to find causal relationships. 
And compare it with two state-of-the-art works.
To evaluate the efficiency and eﬀectiveness of Shylock, 
we validate it on the synthetic and FMRI datasets. 
% First, we investigate its performance on datasets with different time lags. Then, we evaluate its generalization on datasets with different data sizes. Additionally, it's also applied to the real dataset, FMRI.
we answer the following research questions.
\textbf{RQ1:} How is the performance of Shylock on real datasets?
\textbf{RQ2:} How is the performance of Shylock on datasets with different data sizes?
\textbf{RQ3:} How is the performance of Shylock on datasets with different time lags?   
% \begin{itemize}
%   \item \textbf{RQ1:} How is the performance of Shylock on datasets with different time lags?   
%   \item \textbf{RQ2:} How is the performance of Shylock on datasets with different data sizes?
%   \item \textbf{RQ3:} How is the performance of Shylock on real datasets?
% \end{itemize}

\subsection{Datasets and Metrics}
\label{Datasets_Metrics}
\textbf{Datasets:} In order to answer these questions, we introduce one synthetic  dataset and one real dataset.
\begin{itemize}
  \item \textbf{Synthetic Multivariate time series Datasets (Few-shot):} 
   The synthetic datasets support customized conditions such as the num of variables, the scale of time delay, and the causal relationship. More details are described in \secref{sec:gen_data}.
%   \textbf{STCD\_delay:} The synthetic dataset support customized conditions
%   such as the num of the variables, the scale of time delay and the causal relationship.   

%   \textbf{STCD\_nodelay:} The second synthetic dataset is similar to the STCD\_delay. The only difference between them is that there is no time delay.
%   More details about STCD\_delay and STCD\_nodelay are described in \secref{sec:gen_data}.
  
  \item \textbf{Real Multivariate time series Datasets (Normal):}  
  The second benchmark is the common benchmark, functional magnetic resonance imaging data (FMRI) \cite{friston2009causal}, which is time series measuring the relationships between blood flow and different regions in the brain. 
  We select all the 28 sub-datasets with the node\_num in $\{5,10\}$, the timesteps varies from $\{50-5000\}$ and the range of $d$s is $\{10,12,13,21,33\}$. 
  Each variable in this dataset involves self-causation.
  % with an average of more than 200 timesteps and 5 variables. 
  The time delay between cause and effect is not available in FMRI.
  
%   The third commonly used benchmark \cite{huang2021benchmarking} measures the interactions of Arctic Sea Ice and Atmosphere. It involves 12 variables and 39-year time period MTS data. For easy description, we refer it as Arctic sea ice dataset in the following paper. 
\end{itemize} 
  
\textbf{Metrics:} Performance is evaluated using standard metrics: Structural Hamming Distance (SHD), Recall also called true positive rate (TPR) ($\frac{TP}{TP + FN} $ ),  Precision ($\frac{TP}{TP + FP}$ ) and F1 score ($2 \ast \frac{Precision\ast Recall}{Precision + Recall} $). 
SHD measures the minimum edge operations (insertion, deletion, reversal) required to align the predicted graph with the true graph \cite{tsamardinos2006max}.
Precision, Recall, and F1 scores range from 0 to 1, with higher values indicating better performance. An F1 score of 1 indicates perfect model prediction.
Metrics are computed by comparing the predicted attention matrix $A$ to the formalized matrix derived from the ground truth causal graph.

\subsection{Method for synthesizing data}
\label{sec:gen_data}
\textbf{Why synthetic datasets is needed? }
Since there are few causal datasets available in the real-world and existing synthetic methods either lack the support of sequential relation or lack support for causal relationships. So we propose a generalized method for synthesizing multivariate time series.

\textbf{How is the synthetic dataset generated? }
To generates few-shot multivariate time series with causal relationships, we draw inspiration from \cite{NoTears}. First, a causal graph $\mathcal{G}$ is created as a directed acyclic graph (DAG) with $n$ nodes, represented by an adjacency matrix $M_{DAG}$. 
% The $M_{DAG}$ is generated based on the idea of random graph, 
The matrix is generated as a random graph similar to \cite{NoTears}, with $d$ total causal relationships. Each node has an average degree of $d/n$ (in-degree and out-degree combined). Time delays for causal edges are randomly assigned within a user-defined range.

Secondly, initial time series are assigned to nodes with zero in-degree, indicating no cause. To provide flexibility for "effect" changes and better feature representation, the spline interpolation method using a three-moment equation is employed. Solving this equation yields curve functions, from which initial time series are uniformly sampled.

Thirdly, assign the degree to other nodes. A topological sort ensures nodes are processed only after their causal predecessors are generated. Time series data are computed by weighting the time lag matrix and adding noise. In the node time-series data reasoning, according to the first step in the simulation time lag matrix weighting matrix and noise generated values the simulation time sequence data of linear target node is calculated.

\subsection{Baselines}
For RQ1-3, we evaluate Shylock with two baselines NOTEARS \cite{NoTears} and TCDF \cite{tcdf}. The first work, NOTEARS is a score-based DAG structure learning method for the causal discovery method on data without temporal distribution. The second work, TCDF (Temporal Causal Discovery Framework), is a constraint-based framework for causal discovery and 
make use of information-theoretic measures to determine dependencies between time series.

\subsection{Implementation Details}
We conducted all the experiments on a computer with Windows 10, 32 GB memory, and an Intel(R) Xeon(R) Silver 4114 CPU @ 2.20GHz.
For Shylock, the dilation factor $d_{l}$ is initialized as 4, and the sharing kernel size as 4. Initially, the non-diagonal elements of the attention matrix $A$ are initialized to 1.
For TCDF, to make it more suitable for few-shot MTS, the kernel size is setted as 4. 
Other unlisted settings follow the settings of TCDF and NOTEARS.

% For each run, we record their performance, 

\begin{figure}
  \centering
  \subfigure[]{
    \label{subfig:timelag} 
  % \subfigure[F1 scores of the three methods on synthetic dataset]{
  %   \label{subfig:timelag} 
    \includegraphics[scale=0.3]{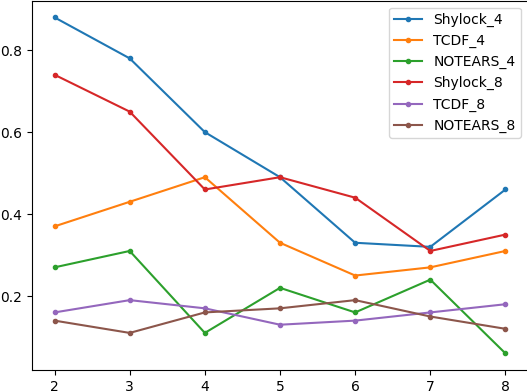}}
    \subfigure[]{
    \label{subfig:datasize} 
  % \subfigure[F1 scores of the three methods on different data\_size of multivariate time series datasets.]{
  %   \label{subfig:datasize} 
    \includegraphics[scale=0.3]{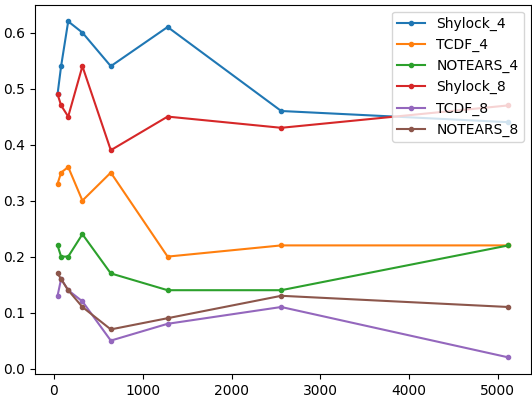}}
  \caption{F1 scores on the synthetic datasets. The chart name consists of two qualifiers. The first qualifier describes the method, the second describes node\_num $n$.}
  \label{timelag_datasize} 
\end{figure}

% \begin{figure}[h]
%   \centering
%   \includegraphics[scale=0.6]{timelag.png}
%   \caption{F1 scores of the three methods on synthetic dataset. Dataset names consist of two qualifiers. The first qualifier describes the method, the second describes node\_num $n$.}
%   \label{timelag}
% \end{figure} 

\subsection{Result Analysis}
\label{Ana_RQ1}

\textbf{RQ1:} In order to assess the ability of Shylock to solve the causal relationship on normal data-set, we compare it with the state-of-the art work TCDF and NOTEARS on FMRI.
% and Arctic sea ice dataset. 

\begin{table}[h]
\centering
\caption{The average results on FMRI of three methods.}
\begin{tabular}{llllll}
\\ \hline
model   & SHD   & avg\_SHD & Precision & Recall & f1   \\ \hline
Shylock & 16.04  & 0.58     & 0.91      & 0.50   & 0.64 \\
TCDF    & 16.79  & 0.75     & 0.59      & 0.56   & 0.57 \\
NOTEARS & 328.39 & 8.83         & 0.77      & 0.1    & 0.18 \\ \hline
\end{tabular}
\label{FMRI_Shylock_TCDF_avg}
\end{table}
We report all the evaluation results on \tabref{FMRI_Shylock_TCDF_NOTEARS} in Appendix. 
% The fist column represents the $i_{th}$ datast in FMRI. The first row represent the evaluation metrics. The second row represents the evaluation metrics in each model. 
For easy description, we report the average results of the evaluation metrics on these 28 sub-datasets in FMRI, shown as \tabref{FMRI_Shylock_TCDF_avg}. 
% We also show the F1 score of all these 28 sub-datasets on \figref{FMRI_3method}. The number on the x-axis represents the number of $i_{th}$ sub-dataset in FMRI and  correspond the first column in \tabref{FMRI_Shylock_TCDF_NOTEARS}.
We report the evaluation results of these three methods with five repetitive times. For Shylock, each run with different initial value in attention matrix. 
Additionally, to fully and fairly evaluate their performance, we introduce $avg\_SHD = SHD/ d$ which reflects the average distance on each causal relationship. For example, there are two datasets where $dataset_1$ involves 100 causal edges, SHD=10, and the other $dataset_2$ involves 10 causal edges, SHD=2. Though the SHD of $dataset_2$ is low, the performance of $dataset_1$ is more satisfactory. 

NOTEARS performs the worst on the FMRI dataset for self-causation is not allowed. Although TCDF performs well evaluated by Recall and performs almost the same as Shylock on SHD and avg\_SHD, it also has a small Precision rate, which makes the overall F1 performance weaker than Shylock. It tends to find false positive causal relationships.

Shylock still performs the best among the three method. 
The F1 score can even achieves 1 in some sub-datasets reflecting it tends matain low FP. 
The Shylock can achieve a balanced performance of Precision and Recall, making the output results of the algorithm more confident.
% It also shows Shylock is effective for finding self-causation. 
Even in the special case in $FMRI\_50\_200\_111\_50$, Shylock still performs the best on the $Precision, Recall, F1$.

% \begin{minipage}{\textwidth}

% \begin{minipage}[t]{0.2\textwidth}
% \makeatletter\def\@captype{table}

% % \caption{The average results on FMRI of three methods.}
% \begin{tabular}{llllll}
% \\ \hline
% model   & SHD   & avg\_SHD & Precision & Recall & f1   \\ \hline
% Shylock & 16.04  & 0.58     & 0.91      & 0.50   & 0.64 \\
% TCDF    & 16.79  & 0.75     & 0.59      & 0.56   & 0.57 \\
% NOTEARS & 328.39 & 8.83         & 0.77      & 0.1    & 0.18 \\ \hline
% \end{tabular}
% \captionof{table}{The average results on FMRI of three methods.}
% \label{FMRI_Shylock_TCDF_avg}

% \end{minipage}

% \begin{minipage}[t]{0.2\textwidth}
% \centering

% \begin{tabular}{llllll}
% \\ \hline
% model   & SHD   & avg\_SHD & Precision & Recall & f1   \\ \hline
% Shylock & 23.61  & 0.82     & 0.58      & 0.53   & 0.55 \\
% TCDF    & 20.07  & 1.17     & 0.32      & 0.37   & 0.34 \\
% NOTEARS & 337.61 & 8.94     & 0.78      & 0.13   & 0.22 \\ \hline
% \end{tabular}
% \captionof{table}{The average results on sampled FMRI of three methods.}
% \label{sample_FMRI_Shylock_TCDF_avg}

% \end{minipage}
% \end{minipage}{\textwidth}

\textbf{RQ2:} To address RQ2, We compare Shylock with TCDF and NoTears for causal discovery and evaluate their performance on the same two benchmarks. The first synthetic dataset is settled with node\_num $n \in\{4, 8\}$, sample\_num $sm \in\{40, 80, 160, 320, 640, 1280, 2560,5120\}$ and time delay $td \in\{2, 3, 4, 5, 6, 7, 8\}$. The second dataset formed by sampling data from the FMRI data set. For each sub-dataset in FMRI, we sampled the last 40 timesteps from the original dataset. 

The experimental results on synthetic dataset are shown in \figref{subfig:datasize}. We can observe that under different sampling quantities, our method still achieves state-of-the-art results, which proves that Shylock not only has superior performance on few-shot datasets, but also has good results when generalized to general datasets.At the same time, it can be seen from the experimental results that with the increase of the number of samples, our method shows better performance.

Results on sampled FMRI dataset are shown in \tabref{sample_FMRI_Shylock_TCDF_NOTEARS} in Section 6 and \tabref{sample_FMRI_Shylock_TCDF_avg}. We can observe that our method Shylock still achieves state-of-the-art results, albeit with a slight drop in prediction accuracy. It proves even on the real few-shot MTS data, Shylock can work well on them.

\begin{table}[h]
\centering
\caption{The average results on sampled FMRI of three methods.}
\begin{tabular}{llllll}
\\ \hline
model   & SHD   & avg\_SHD & Precision & Recall & f1   \\ \hline
Shylock & 23.61  & 0.82     & 0.58      & 0.53   & 0.55 \\
TCDF    & 20.07  & 1.17     & 0.32      & 0.37   & 0.34 \\
NOTEARS & 337.61 & 8.94     & 0.78      & 0.13   & 0.22 \\ \hline
\end{tabular}
\label{sample_FMRI_Shylock_TCDF_avg}
\end{table}

\textbf{RQ3:}  \figref{subfig:timelag} compares Shylock, TCDF, and NOTEARS on synthetic datasets with variables $n \in\{4, 8\}$ and time delay $td \in\{2, 3, 4, 5, 6, 7, 8\}$, while other parameters follow Section \secref{Datasets_Metrics}. Each dataset includes 1 or 2 causes($d \in \{1, 2\}$). 
Shylock consistently achieves state-of-the-art F1 scores across nearly all cases as shown in \figref{subfig:timelag}. 
As time lag increases, noise influences make fitting more challenging, reducing performance. Despite this, Shylock maintains superior results, demonstrating robust causal discovery capabilities.
% \lsadd{It  is greatly affected by the size of time lag. As the time lag increases, the effect becomes worse. More details about reflection}

% Shylock performs better than other methods on few-shot multivariate time series. 
TCDF performs worse for two reasons. Firstly, there are only small amount of data available which can not help TCDF well learn features, for its parameters is exponentially more than the amount of data itself.
Secondly, there is a lack of constraints from a global perspective in TCDF. By analyzing the results of TCDF, WE find that there are a large number of causal loops in the results, most of which are the mutual causal relationship between two nodes for each sub-networks in TCDF can only verify the causal relationship.
NOTEARS performs similar in STCD\_nodelay and STCD\_delay for it cannot make use of the "time" features.  But it can well eliminate cyclic causal relationships. 

Benefiting from global causal constraints and shared parameter kernel design, Shylock enables the model to perform better on few-shot datasets with time lag.

% \subsection{Result Analysis: RQ2}

% \begin{figure}[h]
%   \centering
%   \includegraphics[scale=0.6]{FMRI_3method.png}
%   \caption{F1 scores of the three methods on FMRI.}
%   \label{FMRI_3method}
% \end{figure} 

% Combining the results shown on \tabref{FMRI_Shylock_TCDF_NOTEARS} and \tabref{FMRI_Shylock_TCDF_avg}, we can see that the node\_num $n$ and the length of timesteps $t$ have an important impact on the performance of Shylock and TCDF. 

% \textbf{Fitting ability:} To further compare the fitting ability of Shylock, we further
% compare it with TCDF on STCD with $td \in \{1-4\}$, $d=2$. For fairness, we don't compare NoTears for it is not a deep-learning-based method.
% Then put 70\% of the 
% data in the training set and 30\% in the testing set. Each group of experiments 
% was run on different random seeds, and the average results are displayed to avoid
% the accumulation of errors. As \figref{fit_ability1} shows, compared with TCDF, 
% the stability of Shylock on small data sets is significantly improved. 
% As for the sample num increase, it even performs almost the same as TCDF. We guess 
% it is because TCDF has a large parameter space that can well model the representation
% with large number of MTS. 

% \begin{figure}[h]
%   \centering
%   \includegraphics[scale=0.8]{fit_ability.png}
%   \caption{Comparison of fitting ability between Shylock and TCDF on datasets of different sizes.}
%   \label{fit_ability1}
% \end{figure} 

\subsection{Fitting ability Analysis} \label{fit_ability1}
% \subsection{parameter analysis}'
% To further evaluate the fitting ability of Shylock, we also perform fitting ability analysis.
% Assuming that there are $n$ variables in the MTS dataset, the convolution kernel is set to $n*f$, where $f$ is length of the receptive field. Most neural network based methods constructs $n$ sub-convolution neural networks where $i_{th}$ sub-convolution neural network is used to model the feature representation of $i_{th}$ variable, following the principle of Granger causality \cite{chen2004analyzing}. Then it will combine all these sub-convolution neural networks together to realize the extraction of time series features of all data items, and determine the causal relationship between data items based on the time series features. 
% Although the scheme of multiple sub-convolution neural networks can Improve the fitting ability of the model, but also greatly increase the number of parameters the model has.
To assess the fitting ability of Shylock, we performed a fitting analysis. Assuming an MTS dataset with $n$ variables, the convolution kernel was set to $n \cdot f$, where $f$ represents the receptive field length. Most neural network-based approaches construct $n$ sub-convolutional neural networks, each modeling a specific variable following the principle of Granger causality \cite{chen2004analyzing}. These sub-networks collectively extract time series features and identify causal relationships.

While using multiple sub-networks improves fitting ability, it significantly increases model complexity. For instance, TCDF \cite{tcdf} reduces convolution parameters compared to other methods, but each sub-network still contains at least $O(nkt)$ parameters, with the overall model having $O(nnkt)$ parameters, often approaching or exceeding the input data size for time series scenarios.
% \begin{figure}[h]
%   \centering
%   \includegraphics[scale=0.8]{loss_value.png}
%   \caption{The change of the loss value of the TCDF as the number of samples increases.}
%   \label{loss_value}
% \end{figure} 

The parameter count of TCDF was validated using synthetic datasets, with 70\% for training and 30\% for testing. Experiments were repeated with varying sample sizes from 30 to 4000 across different random seeds, and average results were computed to minimize error accumulation. TCDF employs the Mean Squared Error (MSE) loss function, defined as:
\begin{equation}
  MSEloss = \frac{1}{n} \sum_{i=0}^{n} (y_i - f(x_i| W))^{2}
\end{equation}

\begin{figure}[h]
  \centering
  \includegraphics[scale=0.4]{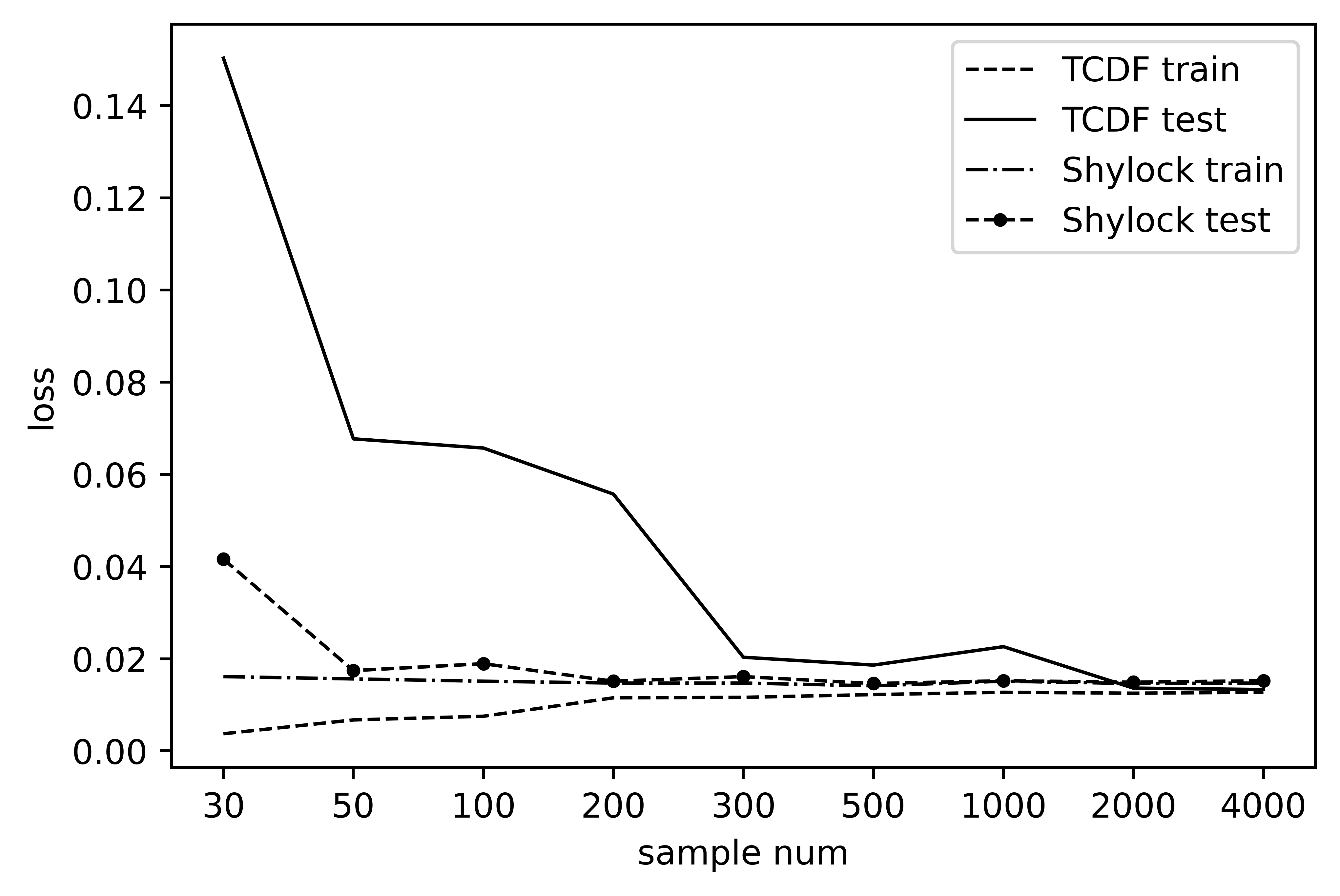}
  \caption{Comparison of fitting ability between Shylock and TCDF on datasets of different sizes.}
  \label{fit_ability1}
\end{figure}

This function effectively models the trend differences between predicted and actual values. As shown in \figref{fit_ability1}, when sample sizes are small, the loss gap between training and testing sets is large, indicating significant prediction deviations and overfitting during training. As sample size increases, this gap narrows, demonstrating TCDF’s limited suitability for few-shot MTS data.

For fairness, NoTears was excluded as it is not a deep learning method. In contrast, Shylock shows greater stability with small datasets, and as sample size increases, its performance approaches TCDF. This suggests Shylock effectively handles both few-shot and larger MTS datasets. TCDF's failure to model few-shot data likely stems from its large parameter space, which supports better modeling of extensive MTS data but leads to overfitting with limited samples.

\subsection{Case study}
\begin{figure}
  \centering
  \subfigure[attention normalized by softmax]{
    \label{subfig:attention_softmax} 
    \includegraphics[scale=0.55]{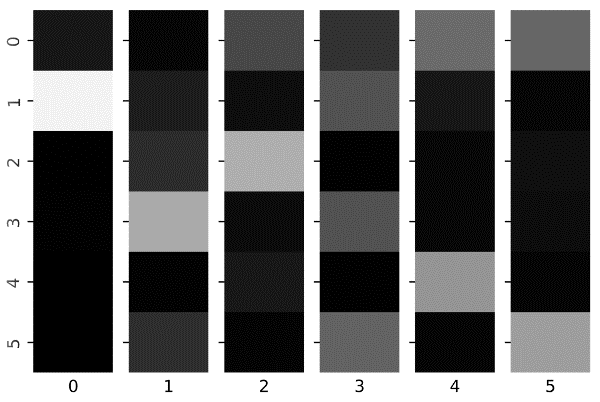}}
  % \hspace{0.5in} % 两图片之间的距离
  % \subfigure[Attention visualization of the dataset.]
  \subfigure[Attention visualization of the dataset.]{
    \label{subfig:visualized_attention_matriix} 
    \includegraphics[scale=0.13]{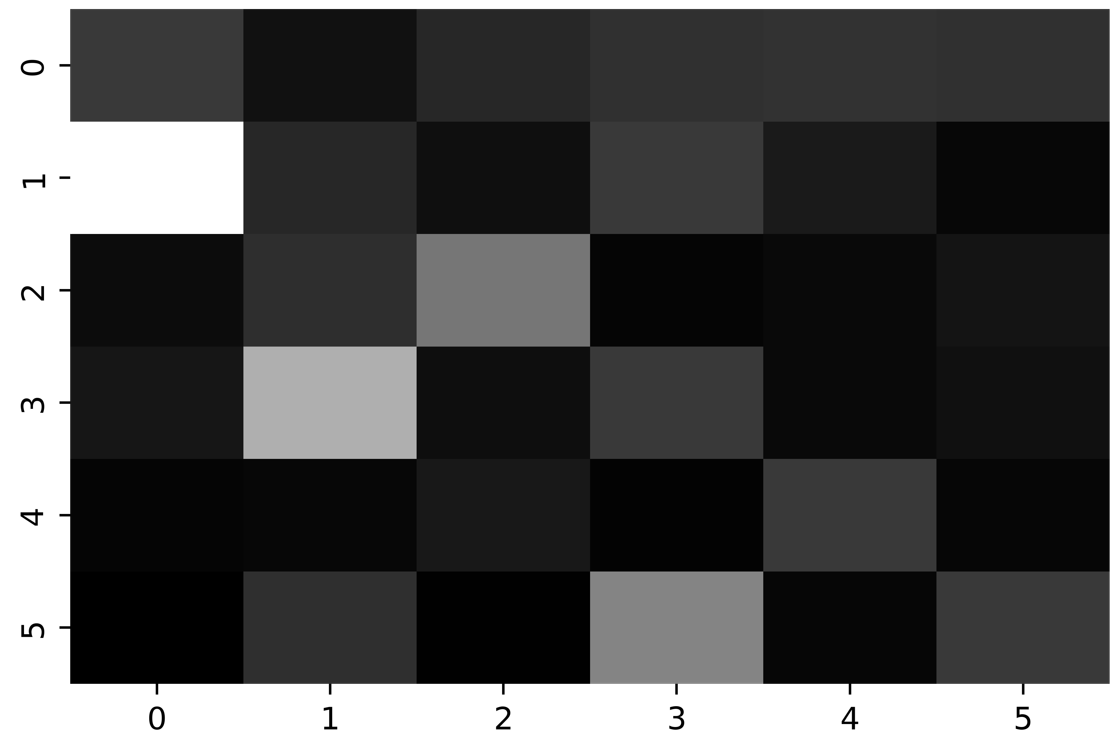}}
  \caption{}
  \label{attention_visualized_attention_matriix} 
\end{figure}

Finally, we randomly select a dataset to illustrate the role of the attention matrix in Shylock.
As shown in \figref{data_relation}, the dataset consists of $n=6$ time series over $t=40$ timesteps, with a time delay $td=4$ and degree $d=6$.
The attention matrix aids in obtaining an intermediate representation of input data. Since Shylock generates sub-networks for each time series, the visualized attention matrix in \figref{subfig:attention_softmax} reveals six sub-graphs, where brighter colors indicate higher weights. The x-axis represents potential "effect" nodes, while the y-axis represents "cause" nodes.

% \begin{figure}[H]
%   \centering
%   \subfigure[The visualized corresponding channels of the attention matrix after the Softmax operation is employed]{
%     \label{subfig:attention_softmax} 
%     \includegraphics[scale=0.4]{attention_softmax}}
  
%   \hspace{0.4in} % 两图片之间的距离
  
%   \subfigure[Attention visualization of the dataset.]{
%     \label{subfig:visualized_attention_matriix} 
%     \includegraphics[scale=0.35]{visualized_attention_matriix}}
%   \caption{Attention matrix}
%   \label{attention_visualized_attention_matriix} 
% \end{figure}

% \begin{figure}[h]
%   \centering
%   \includegraphics[scale=1]{attention_softmax.png}
%   \caption{The visualized corresponding channels of the attention matrix after the Softmax operation is employed.}
%   \label{attention_softmax}
% \end{figure} 
% % The Softmax operation is employed to recalibrate the attention weight of the corresponding channels

% \begin{figure}[h]
%   \centering
%   \includegraphics[scale=0.25]{visualized_attention_matriix.png}
%   \caption{Attention visualization of the dataset.}
%   \label{visualized_attention_matriix}
% \end{figure} 

In \figref{subfig:visualized_attention_matriix}, the (0,1) region is notably bright, indicating a strong causal relationship between $0 \to 1$ where node 0 is the cause and node 1 is the effect.
The attention matrix highlights key causal relationships in the dataset, including $5 \to 3, 3 \to 1$, and $1 \to 0$, as reflected by their bright colors and high weights. Other unrelated edges have smaller weights, demonstrating Shylock is able to accurately capture relevant causal relationships aligned with the ground truth.

\section{Related work}
A range of approaches to causal discovery over time series has been proposed. They can be classified into the following classes.

% \subsection{graphical criteria methods}  
% This kind of method relies on a non-trivial combination of probability axioms to find causal relationships over multivariate data.
% The first line is graphical criteria methods. They rely on a non-trivial combination of probability axioms to find causal relationships over multivariate data. These methods can be further classified into two lines: Constraint-based methods and Score-based methods. 

\textbf{Constraint-based methods}
\cite{spirtes2000causation} \cite{zhang2008completeness} \cite{ogarrio2016hybrid} \cite{triantafillou2016score} are well-known two-phase procedures. 
They first infer the causal relationships by the conditional dependencies imprinted in the data and then search for a DAG that entails all (and only) of these dependencies. 
These methods do not necessarily provide complete causal information because they output (independence) equivalence classes, i.e., a set of causal structures satisfying the same conditional dependencies. One representative research, PCMCI, declared that Including more variables makes an analysis more credible regarding a causal interpretation but may lead to more side effects (e.g., leading to smaller effect sizes). 
It proposed to first perform a condition selection stage to remove irrelevant variables and a conditional independence test designed for highly interdependent time series. Actually, it will introduce high noise for it has to perform a huge amount of conditional independence tests.

\textbf{Score-based methods} 
\cite{friedman1997learning} \cite{friedman2013bayesian} \cite{friedman2013learning} \cite{zheng2018dags}, treat the causal graph as Bayesian networks. They use scoring metrics to evaluate the goodness-of-fit of the learned causal relationships and enforce the method of learning relationships toward high scores. But Score-based methods need to search high-dimension space to find the optimal result which has been demonstrated as NP-complete \cite{cg}. One of the most well-known methods, NOTEARS \cite{NoTears}, transforms the search problem into a purely continuous optimization problem to avoid the NP-complete problem. But it still cannot model the time delay.

% \subsection{Deep learning-based methods} 
% Benefiting from the development of deep learning techniques, some advanced deep learning-based methods have been proposed which can be classified into two classes. 
\textbf{Deep learning-based methods} 
% \textbf{Encoder-Decoder based methods} 
\cite{DBLP:conf/iclr/Shang0B21} \cite{yu2019dag} \cite{ng2022masked} \cite{ng2019graph} \cite{gao2021dag} aim to obtain an intermediate representation that can be used to represent the characteristics of the data in a certain time window. 
It can be used for feature extraction of time series in a scenario with a huge amount of data. However, it cannot learn the feature well with a small amount of data and some cannot well model the time delay. In extreme cases, there may even be cases where the number of parameters in the model is exponentially larger than the actual amount of data. Then, some methods like \cite{tcdf} \cite{zheng2020learning} use Neural Networks to improve the accuracy of results by replacing the conditional independence test and designing different rules to check its results. This kind of method still faces over-parametrization problems. For example, \cite{tcdf} uses multiple convolutional neural networks to model the causal relationships of each variable, and then proposes PIVM to verify the founded causal relationships. It shares the same shortcoming as Neural Network-based methods. Additionally, it's easy to see cyclic casual relationships. 
% 
% Among these, some methods\cite{ng2019graph, gao2021dag} generalize the causal discovery work to a graph autoencoder framework for the causal graph is itself a graph structure. For example, \cite{ng2019graph} presents a graph autoencoder framework to be an alternative generalization of NOTEARS to handle nonlinear causal relations. But it cannot well model the time delay.  

% \section{Discussion}

\section{CONCLUSIONS}
We propose Hybrid constraints-based model, \textbf{Shylock}, a new approach for causal discovery on few-shot and normal multivariate time series. Shylock is developed to better
leverage fewer parameters to learn a better representation of each time series for finding local causal relationships and utilize DAG to impose global constraints to realize information sharing among sub-networks.
The results of the experiment state our model is effective for causal discovery and can provide interpretability by the attention results. We also released a third-party library \textbf{Tcausal}, which has been deployed on earthDataMiner for easy use.

%
% ---- Bibliography ----
%
% BibTeX users should specify bibliography style 'splncs04'.
% References will then be sorted and formatted in the correct style.
% %
% \bibliographystyle{splncs04}
% \bibliography{mybibliography}
\bibliographystyle{splncs04}
\bibliography{TCABIB.bib}

\end{document}